RESEARCH ARTICLE

# Generation of Asset Administration Shell With Large Language Model Agents: Toward Semantic Interoperability in Digital Twins in the Context of Industry 4.0

**YUCHEN XIA, (Member, IEEE), ZHEWEN XIAO, (Member, IEEE), NASSER JAZDI, (Senior Member, IEEE), AND MICHAEL WEYRICH, (Member, IEEE)**

Institute of Industrial Automation and Software Engineering, University of Stuttgart, 70550 Stuttgart, Germany

Corresponding authors: Yuchen Xia (yuchen.xia@ias.uni-stuttgart.de) and Michael Weyrich (michael.weyrich@ias.uni-stuttgart.de)

This work was supported in part by the Stiftung der Deutschen Wirtschaft (SDW); and in part by the Ministry of Science, Research and the Arts of the State of Baden-Wuerttemberg through the Projects of the Exzellenzinitiative II.

**ABSTRACT** This research introduces a novel approach for achieving semantic interoperability in digital twins and assisting the creation of Asset Administration Shell (AAS) as digital twin model within the context of Industry 4.0. The foundational idea of our research is that the communication based on semantics and the generation of meaningful textual data are directly linked, and we posit that these processes are equivalent if the exchanged information can be serialized in text form. Based on this, we construct a "semantic node" data structure in our research to capture the semantic essence of textual data. Then, a system powered by large language models is designed and implemented to process the "semantic node" and generate standardized digital twin models (AAS instance models in the context of Industry 4.0) from raw textual data collected from datasheets describing technical assets. Our evaluation demonstrates an effective generation rate of 62-79%, indicating a substantial proportion of the information from the source text can be translated error-free to the target digital twin instance model with the generative capability of large language models. This result has a direct application in the context of Industry 4.0, and the designed system is implemented as a data model generation tool for reducing the manual effort in creating AAS model by automatically translating unstructured textual data into a standardized AAS model. The generated AAS model can be integrated into AAS-compliant digital twin software for seamless information exchange and communication. In our evaluation, a comparative analysis of different LLMs and an in-depth ablation study of Retrieval-Augmented Generation (RAG) mechanisms provide insights into the effectiveness of LLM systems for interpreting technical concepts and translating data. Our findings emphasize LLMs' capability to automate AAS instance creation and contribute to the broader field of semantic interoperability for digital twins in industrial applications. The prototype implementation and evaluation results are presented on our GitHub Repository: https://github.com/YuchenXia/AASbyLLM.

## I. INTRODUCTION

In a digital twin system, different system parts may represent entities and model data in different ways. These variations can cause interoperability issues, as data exchanged between different parts of the system may not be compatible due to mismatches in modeling and interface specifications. Consequently, communication becomes impossible without a common understanding of data. Typically, resolving this issue involves mapping data points between the various system parts. This mapping process often becomes impractically complex due to its nature of handling combinatorial complexity of O(C(n,2)), approximately $O(n^2)$, where '*n*' represents the different modeling methods.

Industry 4.0 adopts a standardization approach to address this interoperability issue. In this context, the Asset Administration Shell (AAS) [1] is a key concept designed to enable seamless interoperability within a digitalized manufacturing environment. AAS provides a standardized method for modeling information and interfaces associated with a physical asset, such as a machine, a component, or a product. If all digital twin models conform to the standardized AAS specifications, seamless communication and data exchange can be facilitated. This standardization simplifies the communication process, reducing its complexity to a linear scale, O(n), where each model and communication protocol aligns with Industry 4.0 standards [1]

Participants and users within Industry 4.0 often possess vast amounts of data. However, transforming company-specific data into the standardized Asset Administration Shell (AAS) format and manually creating AAS instance models remains a challenging and labor-intensive process. To improve the cost-benefit ratio and encourage broader adoption of the AAS technology stack, it is essential to develop new solutions for automated AAS model creation [2]. Researchers advocate in [3], [4], [5] for the development of new solutions for automated AAS creation based on available heterogeneous engineering information, and [6] poses scientific inquiries regarding how LLMs can be utilized in digital twin engineering. Automating the translation process—from existing vendor-specific information models to standardized AAS models—presents a viable and promising solution to enable interoperability in digital twins with lower costs [3], [4], [5], [6]

### A. KEY FOUNDATIONAL CONCEPTS FOR THE SOLUTION

To handle these problems, we focus on three essential conceptual building blocks: textual data serialization, semantics interpretation, and large language models.

**Textual data serialization**: digital twins' data can typically be converted into textual formats, such as code, structured data exchange files, or plain natural language text. This conversion into text makes it technically feasible to use Large Language Models (LLMs) for processing information within digital twins.

**Semantic interpretation**: the primary focus in information processing and utilization should be based on the meaning conveyed by the data, rather than the raw data itself. Systems can communicate effectively when the data communicated is clearly defined and disambiguously interpreted.

**Large language models**: they are pre-trained to acquire the knowledge patterns to process the conceptual meaning of text. This semantic interpretation capability is evidenced by studies of their internal neuron structures and behaviors [7], [8].

The combination of these three conceptual building blocks suggests that if LLMs can offer capabilities for semantic interpretation, then the information in digital twins can be seamlessly utilized on a semantic level, overcoming challenges such as raw form differences, modeling format misalignments, or vocabulary discrepancies. Reducing these barriers will enhance the interoperability of technical systems, achieving what is referred to as "semantic interoperability", where systems exchange information and operate based on the meaning of the data. Consequently, this will lead to the development of a translation mechanism that streamlines digital information flows in technical information systems.

### B. APPLY LLM TO GENERATE DIGITAL TWIN MODELS IN THE CONTEXT OF INDUSTRY 4.0

Figure 1 articulates the key concepts and the relationships introduced so far. In this paper, we aim to create a theoretical framework that employs LLMs to interpret text-based data and to translate these data according to the target system specification to achieve semantic interoperability in digital twin systems. To concretize and realize the theoretical general framework with an empirical investigation, we use the Asset Administration Shell (AAS) for a case study, which is a standard implementation of digital twins for technical assets under the German initiative "Platform Industry 4.0" [1]. AAS standards facilitate the bi-directional conversion of AAS digital twin models into serialized JSON or XML formats. Based on this foundation, LLMs can directly generate the data elements in these serialized forms and merge the generated model into a digital twin system. An LLM system is designed to process textual data collected from technical data sheets and generate AAS models that digitally represent technical components (sensor, actuator, controller, and network devices) in accordance with the standardized specifications for AAS digital twins.

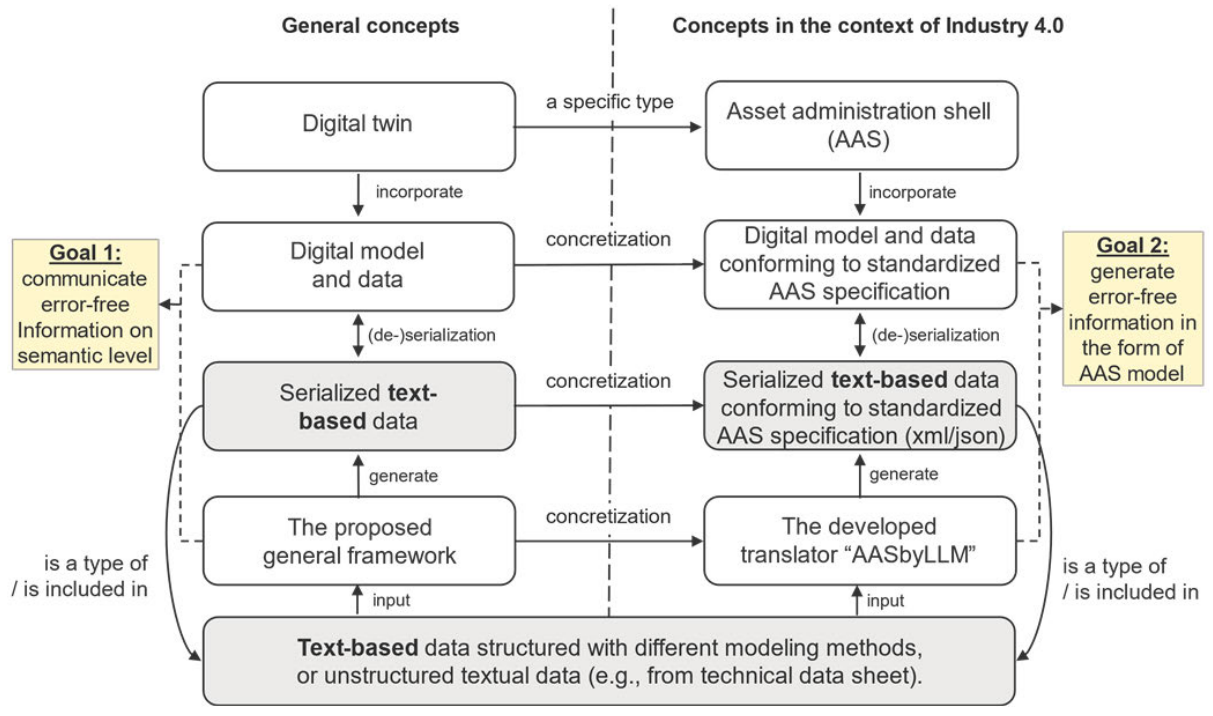

**FIGURE 1. Concept map: This paper's investigation is contextualized within the digital twin concept asset administration Shell in Industry 4.0 and maintains relevance to general digital twins.**

The foundational premise for the investigation of this paper is established on the logical equivalency of two goals:

**TABLE 1.** Literature review on automated AAS model generation.

| Paper | Purpose | Transformation methods | Transformed data type / source format | Semantics in data |
|---|---|---|---|---|
| **Category 1: Using pre-defined rule-based mapping** | | | | |
| [9] | Model Transformation for AAS | Model Transformation Language based on OCL | Software model / UML | Rely on static model semantics in the context of model driven software engineering |
| [10][11] | Interoperable digital twins in IIoT systems | Detailed mapping model for conversion to AAS format | Proprietary information model / UML | Reference on ECLASS dictionary for data definition |
| [12] | AAS creation from heterogeneous Data | Python implementation for extracting and mapping engineering information | Engineering information / PDF, STP, XML, XML/AutomationML, RDF | Mentioned semantic web and semanticID in AAS with ECLASS reference |
| [13] | Mapping between AAS submodels and skill ontology CaSkMan | Rule-based mappings using RML and RDFex | CaSkMan ontology / RDF, OWL | Formal semantic in ontology |
| [14] | Semantic interoperability for AAS-based digital twins using ontologies | Ontology Modeling Language (OML) to map AAS models to OWL | Concepts in Manufacuturing Resource Capability Ontology / OWL, UML | Connecting ontology vocabularies semantics with system models |
| [15] | Interoperability between DTDL and AAS | Mapping between the Digital Twin Definition Language (DTDL) and AAS | DTDL metamodel elements / UML | Semantic annotation in JSON-LD and AAS |
| [16] | Generation of digital twins for information exchange | Formalizes the AAS meta-model in a domain-specific language and an intermediate representation | AAS model / XSD, JSON schema, RDF | Mapping based on semantic equality |
| [17] | Improve semantic discoverability | Establish a set of rules for converting RDF-based models into AAS models | RDF-based models / RDF, SHACL, OWL, SPARQL | RDF as a semantic base for AAS |
| [18] | Generating test cases to verify AAS server implementations | Transform AAS-based plant models into MaRCO Ontology instances | AAS model / UML, RDF, OWL, SPARQL | Semantic interoperability based on ontologies |
| [19] | Provide data in a standardized and semantically described manner | Mapping semantic descriptions from OPC UA information models into the AAS model | OPC UA information models / OPC UA nodes, JSON | Specified semantics in OPC UA standards |
| [20] | Model the semantics of the current operation, status and configurations of assets. | A mapping ruleset to create the semantic AAS as an RDF graph aligned with the RAMI4.0 vocabulary | AAS model / JSON, RDF, SHACL, SPARQL | Using ontological modeling and semantic technologies. |
| [21] | Generating AAS from engineering data | Collect data models into AutomationML and map attributes into the AAS format | Engineering data / AutomationML | Semantics are specified as roles and relations between objects in AutomationML |
| [22] | Interoperability in manufacturing systems for exposing information | Map XML elements to the AddressSpace of an OPC UA | AAS model elements into OPC UA node / XML | Standard semantics in OPC UA model |
| **Category 2: Applying NLP methods, specifically embedding language models** | | | | |
| [2] | Data migration into AAS | Using embedding language model to generate mappings | Technical properties / textual data | Semantic embedding with language models, ECLASS dictionary |
| [23], [24], [25] | Semantic interoperability in heterogeneous AAS | Using trained embedding language model to generate mappings | Technical properties / knowledge graph, textual data | Semantic embedding with language models, ECLASS dictionary |
| [26] | Map technical operational data for building monitoring | Using trained embedding language to automatically map heterogenous data points | Technical properties / Information model from various communication protocols | Semantic embedding with language models, ECLASS dictionary |

***Goal 1:* achieving error-free semantic communication from a source to a target**

*is logically equivalent to*

***Goal 2:* ensuring error-free generation of data required by the target based on the meaning of the source data.**

This investigation serves two primary purposes: First, it demonstrates that our proposed general theory can be technically realized and empirically validated through a representative case study. Second, the software application we implemented translates universal text-based raw data into standardized AAS digital twin models, serving as a tool for automating data translation and enhancing the efficiency of AAS model creation. Finally, the generated AAS model can be integrated into AAS-compliant digital twin software (specifically, AASX Package Explorer) for seamless information exchange.[1]

[1]Prototype at: https://github.com/YuchenXia/AASbyLLM.

### C. PAPER STRUCTURE AND OUR CONTRIBUTION

This paper begins with reviewing existing approaches employed to convert diverse, heterogeneous information into AAS models. It provides a comprehensive review of their applied methods, along with an in-depth discussion of the strengths and weaknesses (c.f. **Table 1**).

Drawing insights and departing from these existing approaches, we introduce our novel theoretical framework that utilizes the large language model to analyze the semantics of textual data in technical information systems and translate them into a self-contained semantic unit called "semantic node" (Section III). We provide a narrative of development in LLM technology to justify why the LLM can be used to process the semantics of textual data (Section IV).

Diving deeper into the practical value of our proposed framework, we conducted a case study using the semantic node and LLM agent system to automate the generation of AAS instance models from raw text obtained from technical data sheets of automation components. Furthermore, we developed a translation software called "AASbyLLM" (cf. **Figure 2**), where users can use this implemented AI tool via web-browser to generate AAS instance model with their own data (Section V). The developed system is comprehensively tested and evaluated (Section VI), and the results are analytically discussed and summarized (Section VII).

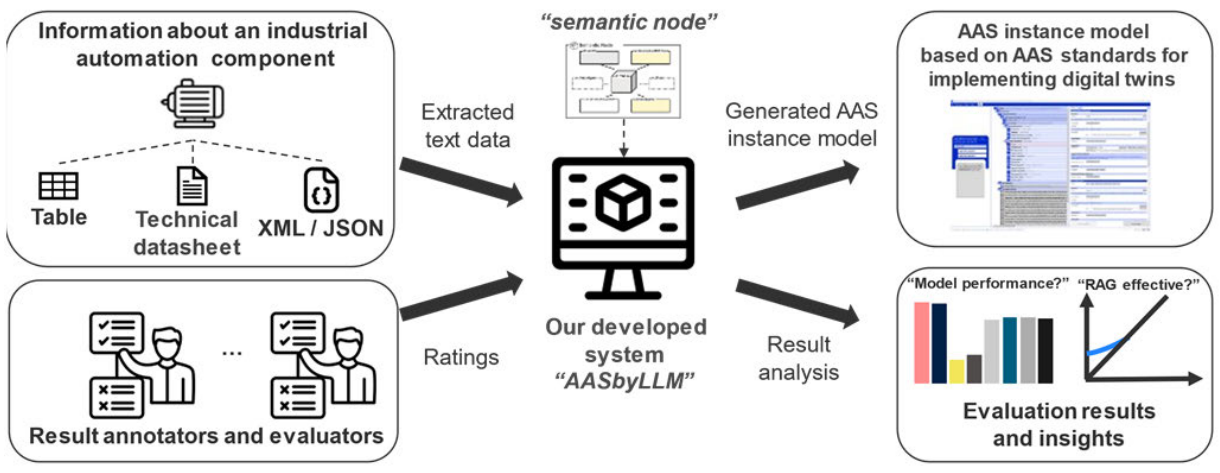

**FIGURE 2. The presented system "AASbyLLM" in this work has two parts: an implemented prototype that delivers the proof-of-concept and an evaluation benchmark based on human feedback.**

The main messages and contribution of the paper are:

- **Theory:** We conceptualize a framework to utilize the large language model to interpret the serialized textual data from digital twins to achieve semantic interoperability and construct a data structure called "semantic node" for capturing the essential semantics during the information processing. A general approach for creating LLM agents system to realize this process is designed and presented.
- **Implementation and Utility:** We show how the designed system can be implemented with an LLM agent system for processing information and generating AAS instance models. By manipulating the "semantic node", an information processing system powered by LLMs can generate informative and correct AAS instance models automatically with 62-79% effective generation rate based on human evaluation, and this indicates that at least a significant proportion of manual creation efforts can be converted into simpler validation efforts, reducing the cost of AAS instance creation. We publish our evaluation dataset and host a web-service for an end-to-end testing of the thoroughness of this approach.
- **Evaluation & Insights**: We performed model-level comparative analysis and ablation study in our evaluation. The results show that both the proprietary model from OpenAI and the open-source models can deliver satisfactory results. Our statistical analysis also revealed the conditional effectiveness and limitations of RAG mechanisms, leading us to propose the "Cheat Sheet Effect of RAG" hypothesis to elucidate these findings. This provides insight into how to design an LLM system effectively.

## II. RELATED WORKS TO AAS INSTANCE CREATION

The Asset Administration Shell (AAS) is a specific type of digital twin that acts as a structured and standardized digital representation of an asset in the context of Industry 4.0. It includes detailed submodels describing the asset's features, properties, and functions, facilitating interoperability across various applications within digitalized manufacturing systems [1].

Based on our literature review, we identified 19 papers closely related to semantic interoperability in the context of Industry 4.0 digital twin modeling and the automated creation AAS instances [2], [9], [10], [11], [12], [13], [14], [15], [16], [17], [18], [19], [20], [21], [22], they are compared and summarized in **Table 1** for better overview. After examining the broad-spectrum methods used to create AAS model and their commonalities, we distilled the essential aspects of these studies into two key categories:

### A. RULE-BASED AUTOMATED CONVERSION METHODS

The most methods focus on the automated conversion between AAS and various information sources through precisely defined and clear-cut mappings and explicitly outlined target and source models. This approach ensures structured data translation but requires significant initial setup for rule definition and managing the precise mappings between semantics between source and target models.

### B. SEMANTIC ENHANCEMENT OF AAS

This emphasizes enhancing data communication through identifiable semantic descriptions of data. It involves assigning clear semantic identifiers from standardized dictionaries or adding detailed descriptions to data elements.

These approaches connect to the concept of 'semantic interoperability', which seeks to overcome the limitations posed by varying data formats and representation styles through identifying the semantic equivalency of data.

## III. GENERAL FRAMEWORK OF THIS WORK

We define a principal proposition of "semantic communication":

> *Effective communication between two systems can be achieved provided that the data being communicated is clearly defined and disambiguously interpreted.*

As listed and summarized in **Table 1**, the most methods [9], [10], [11], [12], [13], [14], [15], [16], [17], [18], [19], [20], [21], [22] utilized the rule-based mapping to bridge the discrepancy of different modeling approach to enable interoperability. Rule-based methods adhere to this proposition and these mappings eliminate ambiguity by clearly specifying how data from one model should be converted to another. However, the drawback of this approach is indeed having these rigorous rules in place. Firstly, this mapping is fixed and constrained by a source model and a target model, making it difficult to adapt to new or evolving data models, these lead to high cost in enumerating and managing of these rules (**inflexibility**). Secondly, there are often nuanced differences between elements of two models, and a rigorous mapping may force subtle changes and lead to semantic shift. This results in information distortion, and an intuitive example is that rule-based translation often falls short in translation tasks in NLP (**distortion**).

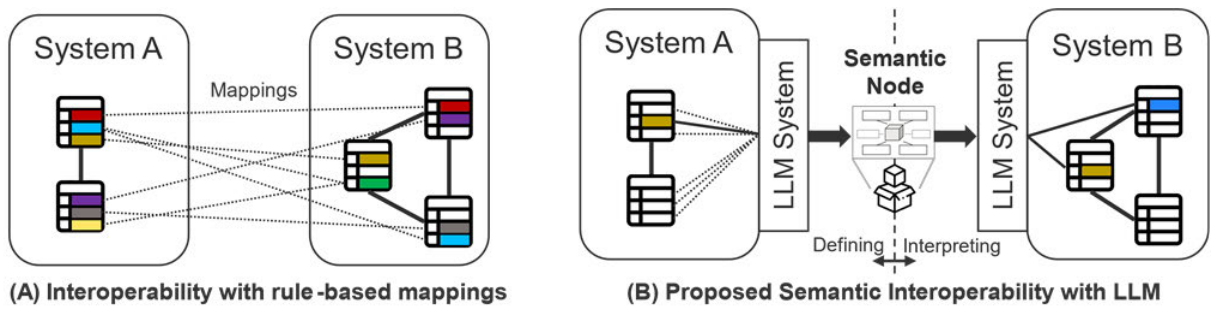

**FIGURE 3.** Interoperability between two systems with different modeling.

The driving idea of this work is to depart from the rule-based transformation model, engage the idea of utilizing semantics, and explore a novel conceptual approach to translate information from one source technical information system to another target system. The theoretical framework of this approach is outlined in **Figure *3***, which emphasizes capturing the meaning of the data being communicated and encapsulates it into an atomic semantic unit called a "semantic node", as illustrated in **Figure 3 (B)**.

This implies that representation form and data format become of less concern, the disparities of diverse domain-specific languages or model specification can be overcome by semantic interpretation capability of LLM, thus achieving semantic interoperability. Following this principle, in this work, the processed data are enriched with meta-data that contribute to articulating the semantic meaning in this intermediate unit called "semantic node". This semantic node contains several text elements used to characterize the semantics of the communicated data: "Name", "Value", "Conceptual definition", "Affordance", "Value type", "Unit", "Source description". The concrete specification is detailed in **Table 2** in Section V-A. "Semantic Node".

## IV. BACKGROUND IN LLM TECHNOLOGY

Building on the proposed theoretical proposition of semantic communication, there remains a concrete question: How can the "data being communicated" be "clearly defined" and "disambiguously interpreted"? To address this, we first provide a summary of the background of LLM technology and introduce the system design. We aim to conclusively answer this question by the end of this section.

### A. LARGE LANGUAGE MODELS INTERPRET SEMANTICS

In 2017, as OpenAI's scientist report in [27] their discovery of "sentiment neuron" – a single semantic unit in a neural network that precisely predicts customers' sentiment in review texts and only response to a specific meaning. This finding encourages further use of training objectives such as masked-language-modeling and next-token-predictions to learn high-quality concept representations from texts. Following this, the invention of transformer architecture [28], its integration into large language models, and the scaling-up of LLMs have led to further new interesting insights: in a LLM, the basic syntactic rules are learned in lower layer while high-level semantic concepts are learned in higher layer [29]; Moreover, 'poly-semantic neurons', which handle multiple unrelated concepts, enable the reuse of neurons in processing semantic through superposition, whereas 'mono-semantic neurons', representing singular meanings, enhance concept differentiation for precise detection and reasoning [8], [7]. These neurons form elemental units of LLM's processing "circuits" [7]. Furthermore, by investigating the correlation between neuron activation and output texts, semantic neurons can be identified [30].[2] These works provide evidence that LLMs possess an inherent structure enabling them to effectively interpret semantics.

### B. EMBEDDING LLM AND GENERATIVE LLM

The evolution of large language models has led to the emergence of two distinct types: embedding LLMs and generative LLMs. **Embedding LLMs** transform text into high-dimensional vector spaces, which can be used for indexing. Semantic relationships can be determined through mathematical operations. For example, semantic similarities between two sentences can be identified using embedding LLM, making them suitable for applications like search engines and recommendation systems. On the other hand, **generative LLMs** specialize in text production. They mostly operate on "causal language modeling" method, where each new token is predicted based on the preceding sequence, a process also known as "auto-regression" or "next-token prediction". This method is particularly effective for content generation, producing contextually coherent text.

### C. PROMPTING OF CAUSAL GENERATIVE LLM

A causal generative LLM operates by focusing on the important preceding tokens, predicting the next token based on probabilistic estimation. From a high-level intuition, a prompt can be viewed as an incomplete text sequence, providing an anchoring context for generating a continuation. The model's response is shaped by how these prompts trigger the LLM's learned patterns, guiding the generation process towards a specific result. This allows for a dynamic interaction with the LLM, where the content generation can be steered through careful prompt design in desired directions to elicit the learned patterns and knowledge in LLM.

### D. LLM AGENTS SYSTEM

Conventionally, *agent-oriented system design* refers to a framework used in software engineering for building complex systems that consist of multiple agents that interact with each other to perform tasks and achieve goals [31]. For LLM applications, this design methodology is especially useful for overcoming challenging tasks by applying task decomposition and modular solution strategies ("divide and conquer").

[2]For a visual example: https://openaipublic.blob.core.windows.net/neuron-explainer/neuron-viewer/index.html.

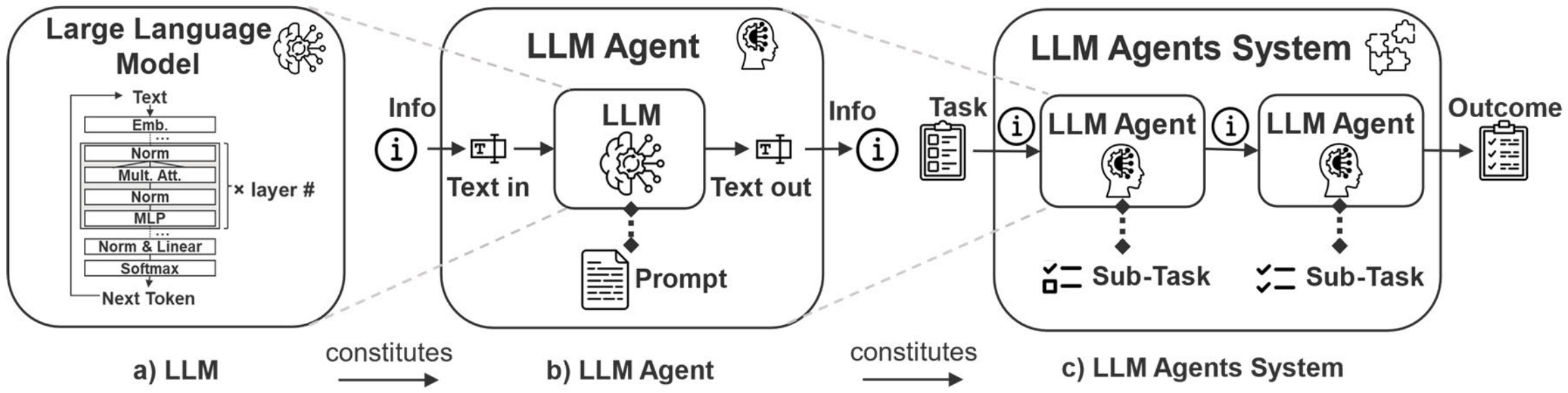

**FIGURE 4.** From single large language model to collaborative LLM agents.

The primary advantage of LLM-agent system is its capability to simplify a complex task by breaking it down into smaller, manageable sub-tasks, each assigned to a specialized agent. A unique aspect of LLM-agents is that their behavior can be defined by the prompt in natural language. In the prompt, defining a specific role is a strategic way to guide and constrain their behavior. In our previous paper [32], we defined a unified structure for specifying prompt for LLM-agent. The method to create a LLM agents system is illustrated in **Figure 4**.

### E. RETRIEVAL-AUGMENTED GENERATION (RAG)

As the name indicates, Retrieval-Augmented Generation (RAG) combines the generative capabilities of LLMs with a retrieval mechanism that queries helpful information from an external knowledge base, such as a database or a vast text corpus. This approach dynamically incorporates retrieved information into the generation process (cf. **Figure 5**). As a result, RAG enhances the LLM's ability to handle tasks that demand task-specific information, especially in scenarios where up-to-date or specialized knowledge is required.

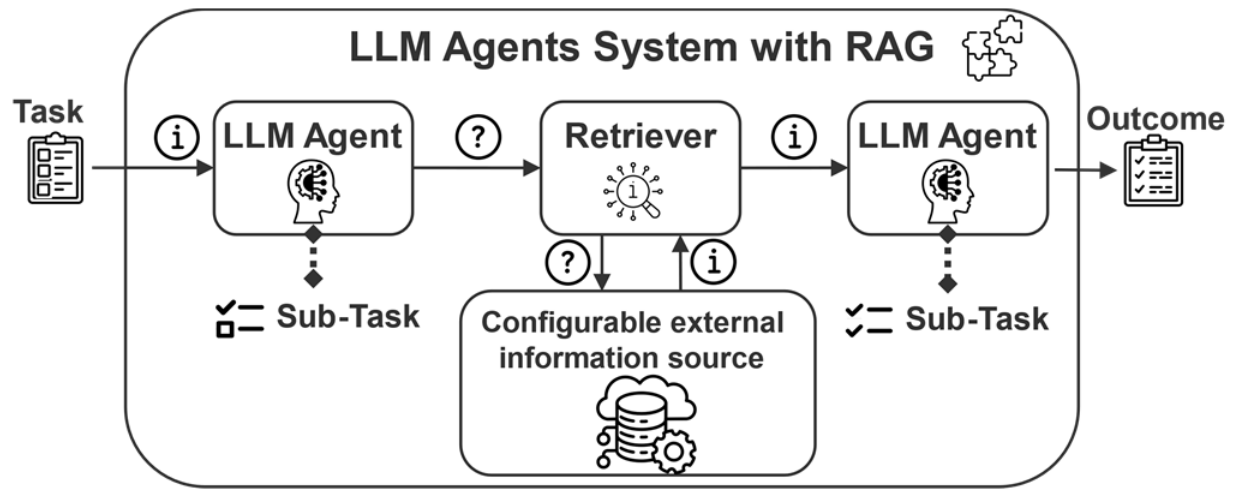

**FIGURE 5.** Integration of Retrieval-Augmented Generation (RAG) in LLM agents system.

### F. CONCLUSION: WHY USE LLM FOR SEMANTIC PROCESSING

Leveraging LLM is a promising approach for ensuring "data being communicated" is "clearly defined" and "disambiguously interpreted". Firstly, LLMs are trained to interpret semantics from extensive training on diverse texts and generate language that aligns with human understanding, and this capability is evidenced by the identification of specialized semantic neurons [30] and reasoning circuit [7]. Secondly, the nuanced interpretation of texts with model internal semantic neuron structures and the ability to distinguish between various meanings enable LLMs to disambiguate terms and phrases effectively. Furthermore, the application of techniques such as text embedding for capturing semantic relationships and generative next-token-prediction for text production allows for precise articulation of concepts in text. Finally, by designing LLM-agent systems and incorporating the mechanism of Retrieval-Augmented Generation (RAG), an information processing pipeline can be created for task-specific applications, for instance, the automation of AAS-instance model creation from raw textual data.

## V. DESIGN AND METHODOLOGY

To handle the semantics of a given piece of information, we propose a data structure called "semantic node", designed to encapsulate various dimensions of meaning. Building on this foundation, we develop the use case and detail the system design.

### A. "SEMANTIC NODE"

The proposed "semantic node" is specified in **Table 2** and visualized in **Figure 10**. The motivation for this is two folds: Firstly, text makes meaning. We want to create an atomic unit that materializes the semantic information into textual elements (c.f. Section III), wherein each text element contributes to defining an aspect of the semantics; Secondly, this data structure facilitates software implementation, and can be independently extracted, modified, evaluated, and converted through the information processing pipeline.

The proposed semantic node is an atomic unit to capture the meaning of a piece of information. This structure is processed by a system constituted of LLM-agents, which synthesizes its elements to enrich the node with specific meanings. The most semantically rich information is considered to be the name, the conceptual definition and the use of data (affordance). The value shall be extracted from the source texts to ensure the nodes' authenticity. Furthermore, the structure includes three optional fields: a source description that enriches background context but requires additional data input; and the value type and unit are not indispensable, because they contribute less to defining the meaning and can often be inferred from other elements (semantic entailment).

**TABLE 2.** Specification of the elements in conceptualized data structure "Semantic Node."

| Semantic node | | |
|---|---|---|
| | **Name** | A concise and specific title assigned to the semantic node. It should be concise yet descriptive enough to convey the essence of the feature at a glance. |
| | **Conceptual definition** | A definition of what the semantic node represents. |
| | **Usage of data (Affordance)** | Describes the potential usage and application of the semantic node. |
| | **Value** | The actual data that is described by the semantic node. |
| | Value type *optional* | The type of data used for representing the actual data. (by default: String) |
| | Unit *optional* | The measurement unit associated with the value. (if applicable) |
| | Source description *optional* | This is an extended explanation that situates the semantic node within the specific context of its source, making the semantic node contextually traceable. (require extra information input) |

This semantic node is a conceptualized structure that serves our theoretical proposition of semantic interoperability. In the next sections, we use this constructed semantic node to design the software application of AAS generation software powered by LLM agents.

## B. USE CASE DESCRIPTION

The designed LLM-based system serves two primary purposes: firstly, for system users, it facilitates the creation of AAS instances by automatically generating data models that describe the technical properties of automation components, thereby reducing manual effort in data migration. Secondly, for research purposes, the LLM produces semantic descriptions for technical data in text, as specified in "semantic node", as lustrated in **Figure 2** at the beginning of this paper. These generated texts are not only integral to the AAS for conveying technical information but also serve as evidence for evaluation. By examining these generated texts, we assess the LLM-system's ability to accurately interpret specialized technical concepts and gauge the overall performance of the LLM in processing and interpreting the semantics of technical data. This dual functionality underscores the system's utility in enhancing semantic interoperability in digital twin implementations and providing a measurable framework to evaluate semantic interpretation capabilities.

**Figure 6** depicts the functions of "AASbyLLM" and the interacting stakeholders in this work. The software provides a web application for users to test the thoroughness of our system and assist us (annotators and evaluators) during the evaluation process. Additionally, the validated high-quality semantic definition generated by the LLM-system can be collected by a library manager and used for enrichment of a dictionary (e.g., ECLASS).

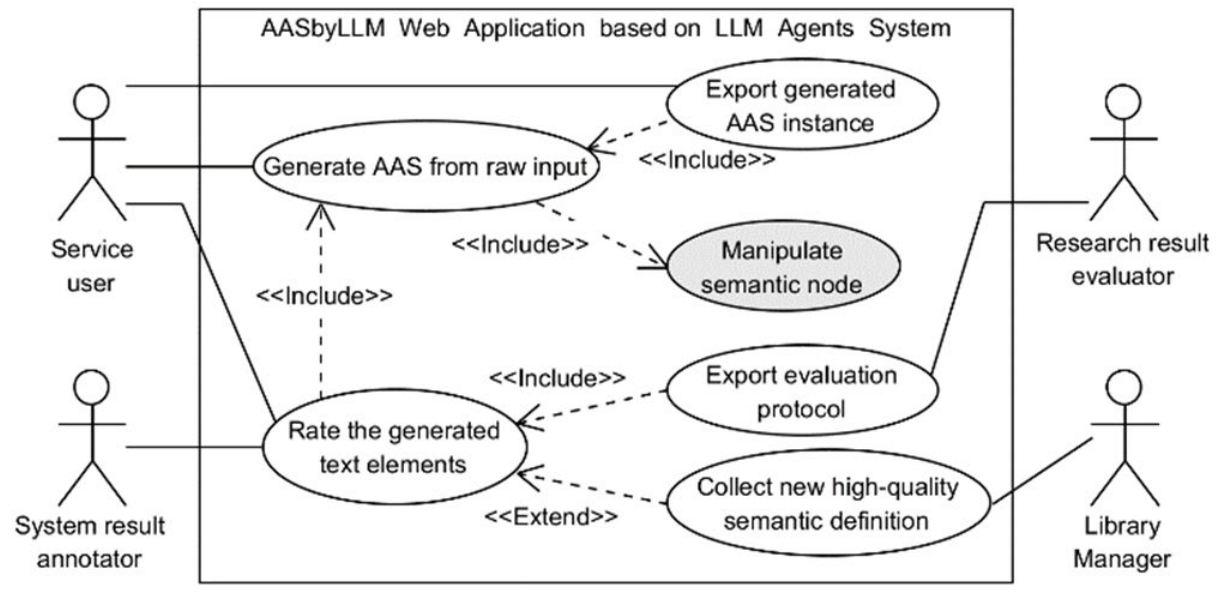

**FIGURE 6.** Use case diagram of the AASbyLLM application.

## C. SYSTEM DESIGN

The system component diagram **Figure 7** presents an overview of the components and the processing pathways involved in the handling of semantic nodes. Its primary objective is to automate the creation of a structured AAS instance model. The process begins with user interaction via a user interface, where raw data comprising technical properties is input. This data then undergoes semantic enrichment and manipulation by the LLM agents in the form of semantic node, eventually producing an AAS model instance based on AAS standard specifications in **Figure 8** [1]. The system components are introduced as follows:

**User Interface:** The user interface serves as the starting point and the user initializes the generation process by submitting the raw input textual data. This input may range from structured data such as spreadsheets or JSON-files to unstructured text such as copy-pasted paragraphs from a technical manual or text from a table of technical properties.

**Retrieval-Augmented Generation (RAG):** After initial input, the RAG sub-system group is engaged, which combines the generative LLMs with information retrieval mechanism from an external domain specific dictionary library. This RAG subsystem consists of several LLM-agents.

**- Identification and Extraction Agent:** In the RAG subsystem. An LLM-agent is designed to identify and extract the name, the value, and an initial definition for a semantic node from the input text. This LLM processes the given input text and initially creates a name, definition, and contextual description for each semantic node as output, enriching the raw data with semantic details in a data structure.

**- Semantic Search Agent:** Following identification and extraction, this system component performs a semantic search using an embedding LLM to find semantically similar entries in the ECLASS dictionary. The search mechanism is based on our previous work [2], where a vectorized embedding index, called "semantic fingerprint", is created for comparison between queried text and each ECLASS dictionary entry. The result is a list of retrieved similar definition entries from ECLASS dictionary.

**- Synthesis Agent:** This step incorporates the results from the semantic search into the generation process. An LLM-agent is prompted to generate a judgment of the relevance of the retrieved entries, accompanied by a short reason in text. The purpose of this step is two folds: firstly, semantic search is based on relationship of semantic similarity, which is a typical proxy metric for search but suboptimal for determining precise relevance, and inappropriate results shall be filtered out; Secondly, in this step, the generated judgement and reason serve as intermediate textual material for considering more nuanced relationships during the whole information analysis process by LLM. By instructing the LLM to judge and reason for each search result, the LLM generates more precise semantic node. After synthesis, a complete semantic node is created based on RAG, ready for AAS model creation. Details about this process are illustrated in our web demo under tab "LLM reasoning details", accessible per link on our GitHub Repository.

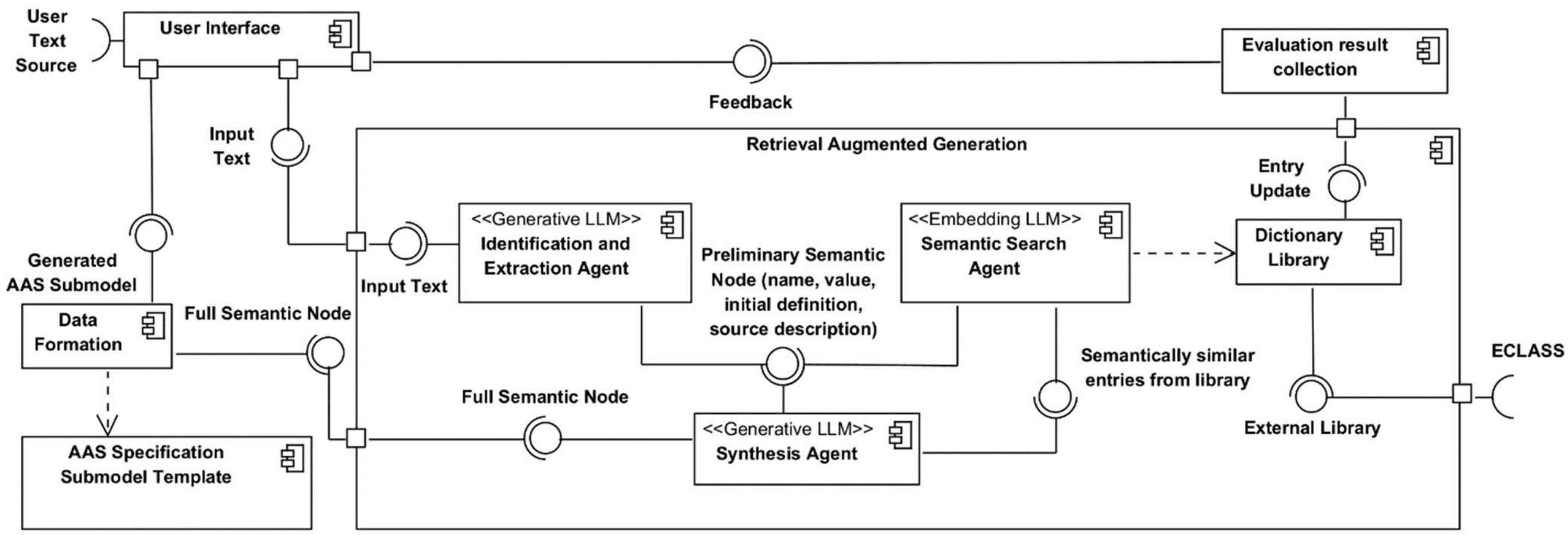

**FIGURE 7.** System components diagram of software "AASbyLLM" for AAS instance model generation with LLM agents system.

**AAS Data Formation:** The system then proceeds to form the AAS model by moving the data from the synthesized semantic nodes into an AAS JSON template. The system leverages established AAS framework to store the generated texts field from semantic node for technical properties of automation components. Specifically, it fills property name in "IdShort" and extracted value in "value", and other fields in "ConceptDescription". Furthermore, we recognize the importance of using identifiable semantics to identify the meaning of data properties and assign a unique "SemanticID" for each newly generated concept definition. This process ensures that the final AAS instance model is in the correct format and adheres to the AAS standard specification, as shown in **Figure 8**.

**User Interface Feedback:** The formed AAS model is then presented back to the user via the UI. Users can inspect the generated AAS model, assess its accuracy and quality, and download the created AAS instance model. Meanwhile, the system collects the user rating.

**Potential Dictionary Library Enhancement:** It would be infeasible for a dictionary to cover all possible meanings in the world. For this specific use case, where the dictionary entries (e.g., ECLASS dictionary library) in some cases may not be perfectly aligned with current properties, more accurate definitions of data property can be generated by the system. Once verified by human experts, these refined definitions can be used for library enrichment.

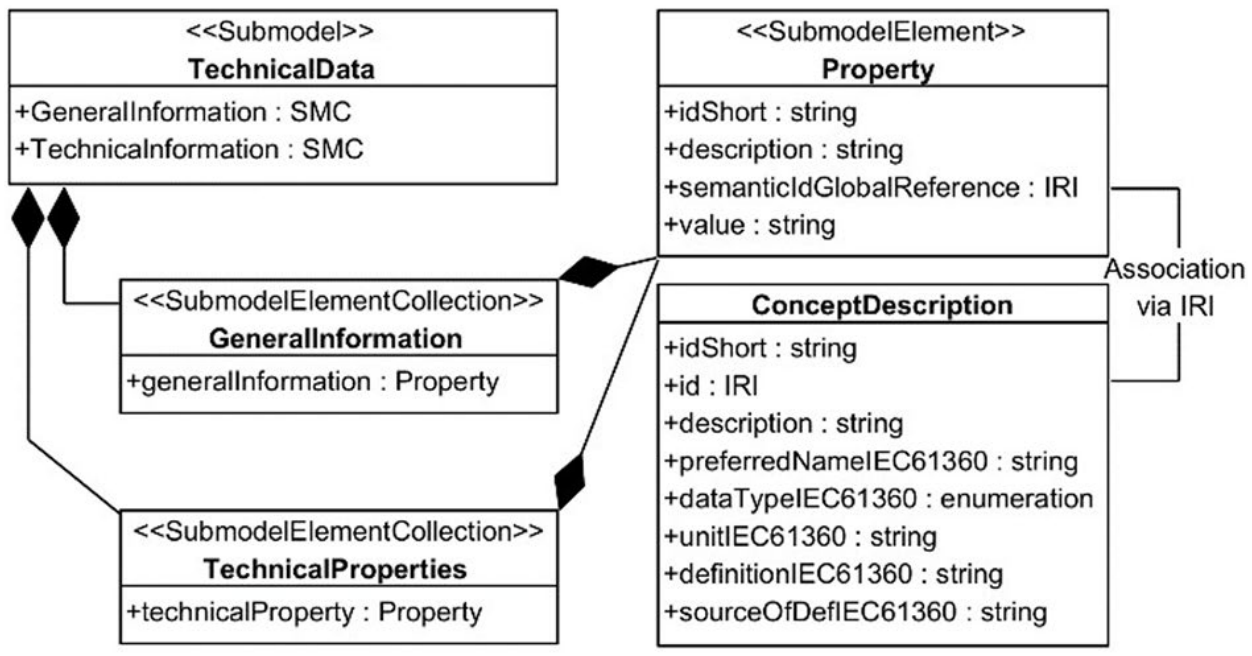

**FIGURE 8.** The AAS model elements generated by the implemented software AASbyLLM; modeling based on [1].

### D. PROMPT DESIGN

The LLM-agent is the elemental function component in the designed system, and it is defined and realized with prompting. We designed a structured prompt template for instructing the behavior of generative LLM-agents. This prompt template is introduced and explained in our previous work based on insights of NLP-research. The essential constitutes and key elements of this structured prompt are introduced in **Figure 9**. More concrete and detailed examples are released in the GitHub Repository.

### E. MODEL SELECTION

We selected four models for comparison, chosen for their similar size and comparable performance. These include the proprietary commercial model *GPT-3.5* and three open-source models deployable on local GPU servers: "*Llama_2_70B_HF*", "*Mixtral_8* × *7B_Instruct_v0.1*", and "*WizardLM_70B_v1.0*".

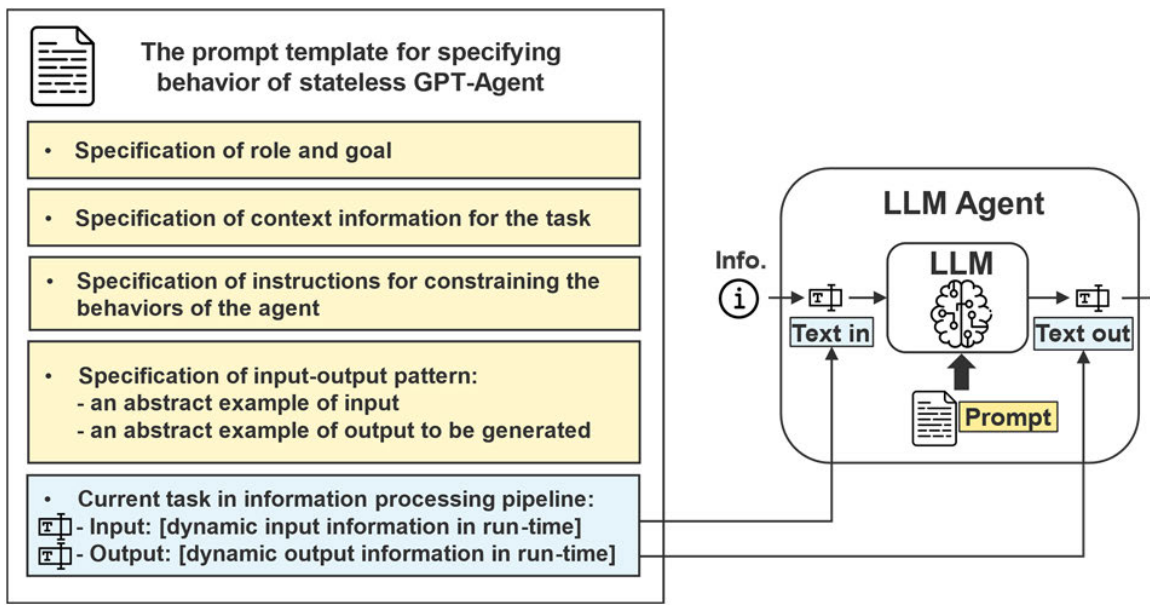

**FIGURE 9.** **The proposed prompt design for creating LLM agent.**

*"Llama_2_70B_HF"* [33] is an autoregressive language model and undergoes supervised fine-tuning (SFT) and reinforcement learning with human feedback (RLHF) post-training to align with human preferences for helpfulness. It serves as the baseline model for our comparison of open-source models.

"*Mixtral_8* × *7B_Instruct*" [34] is a model further developed from "Llama_2_7B" based on the Sparse Mixture of Experts (SMoE) architecture and fine-tuned for following instructions. It is a representative model that utilizes innovative architectural designs to improve performance, enabling it to discern complex patterns and deliver more accurate predictions.

*"WizardLM_70B_v1.0"* [35] is a model further fine-tuned from Llama. Its specialty lies in using synthetical complex instructions (called "Evol-instruct" in [35]) to train a base model solving more difficult tasks. This model exemplifies how further training on a curated dataset can enhance a model. This method is promising because it utilizes automatically generated synthetical data for scalable training of LLMs. This implies that the large language models can train themselves to become more performant.

These models are used to power the designed system. By having a comparison basis and evaluating the outcome, we can gain insights into how the model's capability and system design can have an impact on downstream task results, how effective can our theory be realized and how well the AAS-generation tool performs.

## VI. EVALUATION

Evaluating the meaning expressed by language can be a challenge due to the unlimited combination of texts in natural language, which can result in an immeasurable number of possible expressions. Automatic evaluation is not appropriate. We did a human evaluation benchmark for this work and developed a web-based software to support the evaluation under various configurations. We also formulated proxy metrics to comprehensively assess the system's effectiveness. The evaluation is categorized into three sections: 1) an end-to-end evaluation for practical usability, 2) a comparative analysis of capability of 4 different LLMs, and 3) an ablation study to investigate the importance of integrating external information through Retrieval-Augmented Generation (RAG), with or without utilizing the external ECLASS library.

The evaluation is fundamentally linked to how well humans perceive the generated texts to be **error-free** and **informative**. Though the evaluation leans on human perception, it yields quantifiable results based on statistical principle of large number, indicating the effectiveness of our methodology (pass rate, helpfulness score, element-level rating scores, and inaccuracy rate). Ultimately, the utility of the proposed Asset Administration Shell (AAS) generation system is assessed based on the extent to which it reduces human effort, highlighting the practicality of our AAS generation solution.

### A. EVALUATION SETUPS

The evaluation assesses the ability of the designed LLM-system to interpret technical concepts within an automation system. Technical properties representing the concepts were drawn from 20 automation components, achieving diverse coverage by including 5 sensors, 5 actuators, 5 controllers, and 5 connectivity components, whose technical information is collected from different source. The text results produced by the system provide the primary evidence for its assessment.

In the comparative study, 8 design variations (4 large language models × 2 mechanisms, w./w.o. RAG) of the system were examined. These variations involved four distinct models that have representative characteristics, each tested in two mechanism configurations: one with the integration of the external ECLASS library for Retrieval-Augmented Generation (RAG) and one without it. 7 graduate students with engineering backgrounds with above-average GPA were chosen as annotators to closely resemble the user knowledge profiles. The annotators' individual knowledge level and preference are confounding factors in this experiment. To mitigate potential bias, we randomly allocated the 20 groups of technical properties among 7 annotators and examinate the statistical results calculated from large sample size.

### B. EVALUATION EXECUTION WITH SOFTWARE SUPPORT

The human evaluators are first tasked with reading the technical properties from data sheet of an automation device and making binary statement whether they comprehend the meaning of these technical properties. Then, they determine whether there are errors in the generated 5 data elements ("name", "value", "conceptual definition", "affordance", "value unit"), any inaccuracy in texts or contradiction between their understanding and the generated elements is determined as an error. Then, they are tasked to rate the text quality of the two most semantic-rich elements on a scale 1 to 5: "**conceptual definition**" and "**use of data (affordance)**". Additionally, they evaluate whether the generated texts help them understand the clear meaning of the technical concept if they did not comprehend the meaning of the technical concept at the beginning ("**helpfulness**"). Finally, they determine whether the overall generated semantic node has the quality to be released without human modification and further improvement, this "**overall**" perceived quality is also on a scale 1 to 5. This evaluation process is visualized in **Figure 10**.

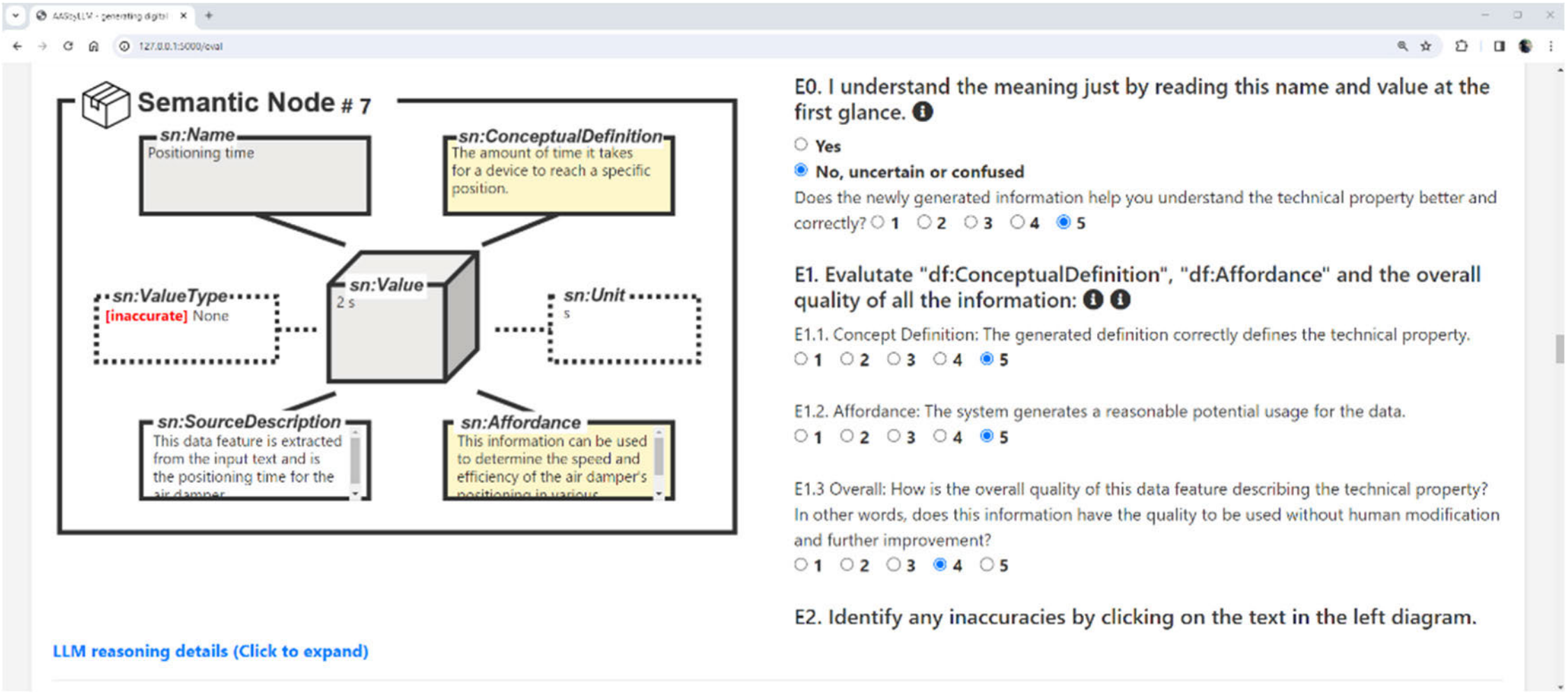

**FIGURE 10.** Screenshot of the application UI used for supporting the annotating and evaluating process.

The evaluation relies on the statistical large number to yield valid results. During the evaluation, a total of 2,259 technical property samples across all the 20 selected automation components by using 4 models and 2 design mechanisms were assessed. Derived from making 10 distinct rating decisions for each individual technical property, this comprehensive assessment resulted in a collection of 22,590 total evaluation decisions. The details of the evaluation are released on our GitHub Repository, including raw data, software web-application, evaluation raw results, calculation spreadsheets, and the analyzed results.

## *C. END-TO-END EVALUATION*

For the end-to-end evaluation of the system designed to automate the generation of Asset Administration Shell (AAS) models using Large Language Models (LLMs), two distinct metrics are proposed: **the effective generation rate**(pass rate) and **helpfulness** (ability to provide implicit knowledge).

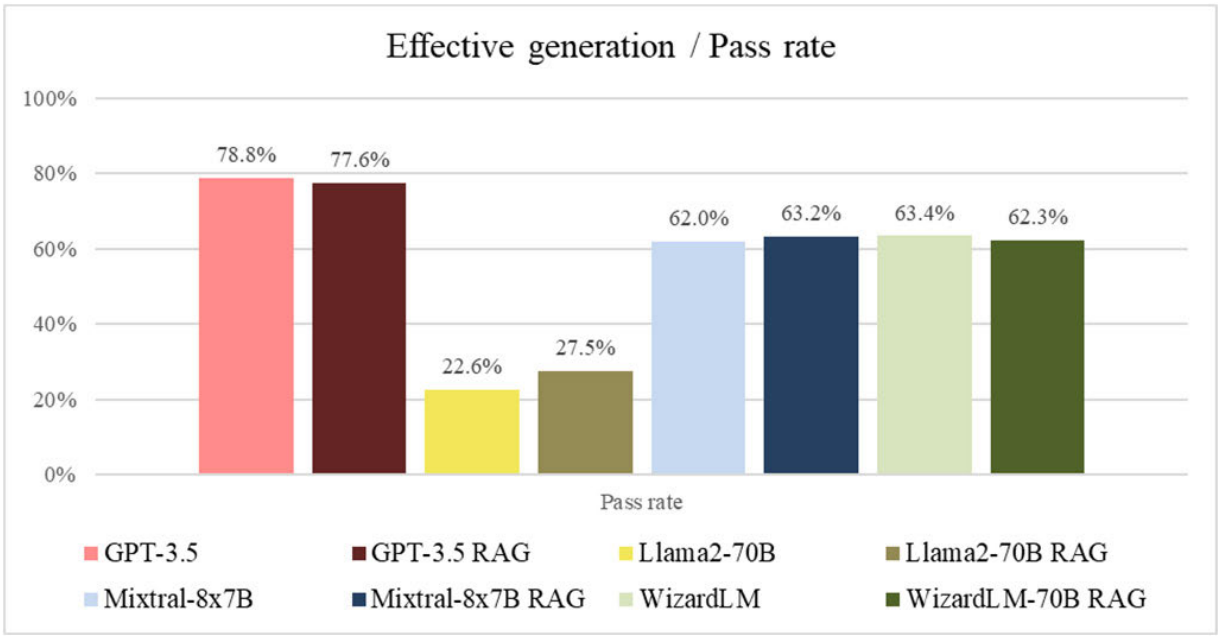

**FIGURE 11.** End-to-end pass rate of the generated instance model under different system configurations, a passed case means that the generated semantic node of a technical property is ready for release without the need for further human modifications.

### 1) EFFECTIVE GENERATION RATE

The end-to-end evaluation aims to measure the capability of the designed LLM-system to generate semantic information, referred to as a "semantic node". This end-to-end evaluation metric is termed "pass rate" or "percentage of effective generation", which measures if the generated result is **error-free** and **informative**. For a sample to achieve the "passed" status (cf. **Figure 11** above):

$$\begin{aligned}\text{Passed}(s) = {} & \text{NoInaccurateText}(s) \\ & \wedge (\text{DefinitionRating}(s) == 5) \\ & \wedge (\text{AffordanceRating}(s) == 5) \\ & \wedge (\text{OverallRating}(s) == 5)\end{aligned}$$

The pass rate is a proxy metric for effectiveness of the system, indicating the readiness of generated information models for practical application without the need for further human modifications.

### 2) HELPFULNESS SCORE

In the technical domain, descriptive texts (e.g., text in a technical datasheet) often assume that the reader possesses prior knowledge of specialized concepts, which can lead to confusion and misunderstandings. This issue of "**implicit knowledge assumption**" becomes prominent when the presented data is inherently complex or incomplete, relying on a foundation of technical knowledge of an agent (machine/human) to correctly interpret the data's meaning.

During human evaluation, the annotator first assesses whether the bare name and value of the technical property alone provide a clear sense making. If not, the system then reveals the generated texts for review. The evaluator then scores the overall helpfulness of these elements in semantic node. Based on their effectiveness in clarifying the meaning and enhancing understanding, the annotator rates on a scale from 1 to 5. Specifically, the annotator is prompt to answer the following question: ''On a scale of 1 to 5, does the newly generated information help you understand the technical property better and correctly?''.

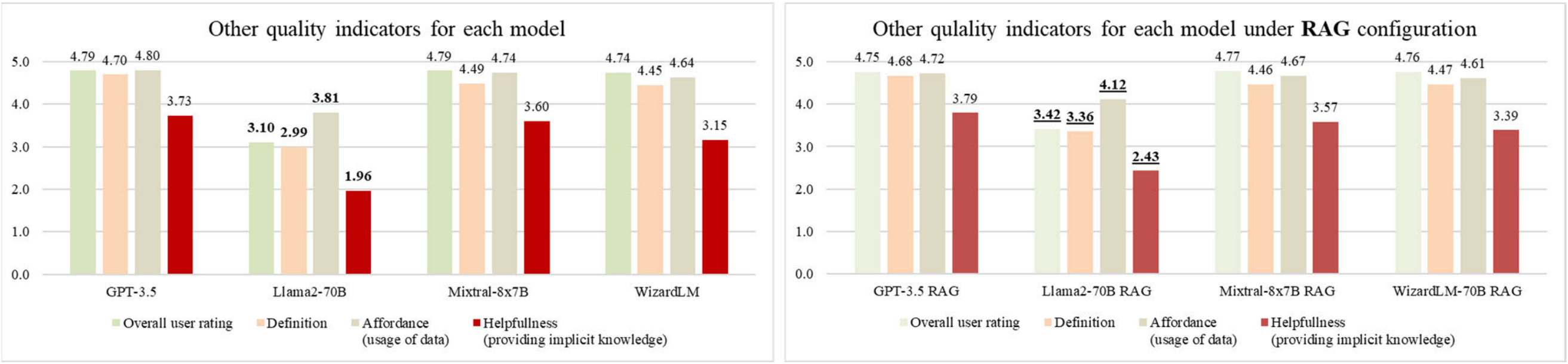

**FIGURE 12.** **Comparative system performance across different configurations with and without RAG mechanism. Each group represents the performance of a system powered by a specific model, with each color indicating a proxy metric for a particular quality indicator. The comparison between the two diagrams demonstrates the effectiveness of the RAG mechanism.**

The resulting so called ''helpfulness score'' proxy metric evaluates the system's ability to provide this implicit knowledge, i.e., to generate texts that aid human users in comprehending the clear meaning of technical properties beyond just name and value. This helpfulness proxy metric is presented in **Figure 12**, together with other detailed metrics (*overall quality, definition text quality, affordance text quality*).

$$\text{Helpfulness Score} = \frac{\sum \text{Rating of each confusing case}}{\#\,\text{confusing cases}}$$

This proxy metric reflects the knowledge capability of an LLM-powered system, specifically, the capability to determine and produce textual data aligned to specialized concepts. From a communications perspective, it indicates that even if the sender makes implicit assumption and simplify the data being communicated to merely a name and value (not loss-free), recipients can still accurately interpret the intended semantics, as long as the recipient has the knowledge to determine the semantics. Additionally, this metric highlights the LLM's ability to add informative explanation and help humans in understanding specialized technical data in scenarios of non-loss-free communication.

#### 3) DETAILED VIEW ON INACCURATE GENERATION

**Figure 13** details the inaccurate generation ratio. The Llama2 based model cannot reliably generate the meaning and usage of the data, as indicated by the ''definition_inaccurate'' and ''affordance_inaccurate'' bars. While the other models rarely generate incorrect texts.

The inaccurate generation ratio in general correlates with the previous evaluation results. RAG only has significant effects for Llama-2 model.

### *D. ABLATION STUDY OF RAG MECHANISM*

Retrieval augmented generation is seen as a promising mechanism to enhance the performance of LLM-system by dynamically incorporating queried data from external knowledge base during the text generation process. To evaluate the RAG mechanism's effect, we conducted a controlled experiment comparing two configurations of our LLM system: one with RAG based on ECLASS-dictionary data definitions, and another where RAG was disabled. The impact of the Retrieval-Augmented Generation (RAG) mechanism on the overall performance of each model varies, revealing distinct patterns. The diagrams in **Figure 12** and **Figure 13** visually demonstrates this effect, comparing the results with RAG and the results without RAG. We summarize the insights as follows:

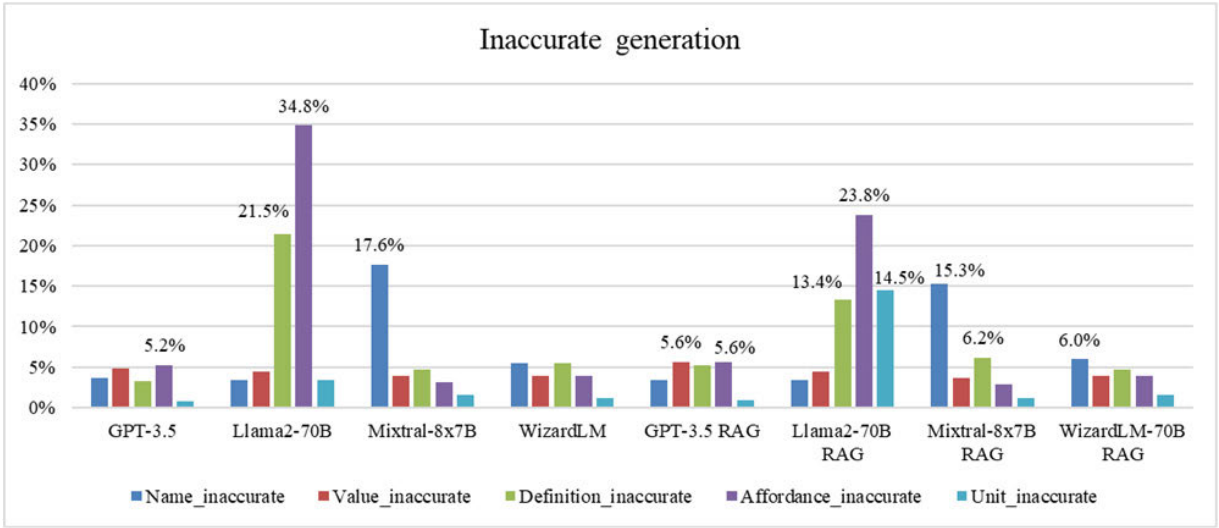

**FIGURE 13.** **Comparative analysis of inaccurate generation rates by different LLMs with and without RAG.**

#### 1) RAG HAS SIGNIFICANT EFFECT FOR WEAKER MODEL

LLAMA-2, which initially had the lowest effective generation rate, shows significant improvement with the introduction of RAG. We calculate the statistical significance with unequal variances t-test (Welch's t-test) for each metric based on two configurations (with RAG or without RAG) with a significance threshold of 0.05. All metrics (helpfulness, overall rating, generated definition rating, affordance rating) indicate significant improvement. The t-test result is released on the GitHub Repository, together with other detailed calculation processes and results during the statistical analysis in a spreadsheet.

#### 2) RAG HAS NO SIGNIFICANT EFFECT FOR STRONGER MODELS

Surprisingly, for the other three stronger models (GPT3.5, Mixtral, WizardLM), the experiment results indicate no significant effect of RAG on enhancing or degrading the qual-ity of the generated texts on all the metrics. This observation suggests that the stronger models may already possess the necessary knowledge to excel in this particular task, making additional information from the RAG mechanism redundant. An explanation is that these models have effectively learned the relevant technical concepts during their training phase, indicating (hypothetically) that a saturation point has been reached where RAG's contribution of new knowledge does not further enhance model performance.

#### 3) IMPLICATIONS FOR USING RAG

An overall intuition that can be drawn from this evaluation is that RAG can enhance the performance of weaker models, but the overall enhancement result may not surpass the performance of fine-tuned stronger models, as illustrated in **Figure 14**.

This dynamic can be metaphorically characterized as "**cheat sheet effect**" of RAG, where weaker language model can be improved with external help, while stronger model does not need to.

While RAG can guide the style and direction of the output by incorporating text from external dictionaries, the semantic similarity search component may also **introduce noise** by fetching **similar yet irrelevant** information from the external database, affecting the final output's relevance and accuracy.

Furthermore, another drawback of the RAG is its **lower efficiency**. The implemented RAG mode statistically consumes about 4.2 times the processing duration, as it needs to process more integrated texts and more numbers of model invocations to generate the final response.

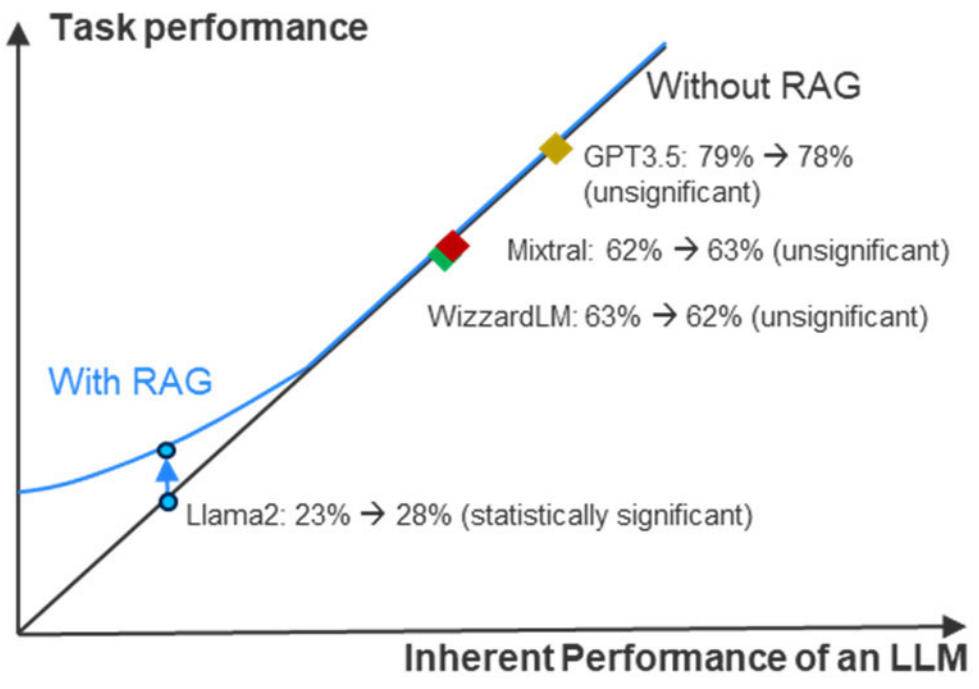

**FIGURE 14. The graphical illustration of the hypothesis how RAG affect the performance of LLM system in technical concepts understanding based on our observed results. The lines are drawn based on assumed interpolation and extrapolation. Without RAG, a LLM only uses its inherent knowledge to execute task, and task performance shall be equal to the performance of the LLM (the 1:1 line); with RAG, significant boosting effect has been observed for weaker model Llama2 in our experiment, while no significant effect has been observed for stronger models.**

From these findings and the assumption that these stronger models should have learned the conceptual meaning of technical concepts, we can formulate a hypothesis based on observed results:

> *The effectiveness of the RAG mechanism is reliant upon its ability to introduce new, previously unknown knowledge to compensate for capabilities that the LLM lacks.*

This hypothesis highlights the importance of carefully considering the effectiveness of the RAG mechanism, particularly in determining when its application is beneficial. It suggests that the decision to employ RAG should be based on an assessment of the gap between the LLM's existing knowledge basis and the required knowledge of the task at hand. Furthermore, this insight can guide the strategic development of RAG datasets, which should be tailored to fill knowledge gaps, thereby maximizing the performance of the LLM system for specific tasks.

## VII. DISCUSSION OF POTENTIAL IMPACTS

In this section, we discuss the implications and potential impact of our work, exploring how it contributes to the research community and the broader field of industrial digitalization.

### A. STREAMLINING DIGITAL TRANSFORMATION WITH LLMs

The generated AAS models can be integrated seamlessly into AAS-compliant digital twin software, facilitating seamless information exchange and communication.

The use of LLM to automate AAS-instances generation provides a technological catalysator for information exchange in manufacturing and digital twin technologies. This approach not only simplifies the creation of standardized digital representations for assets but also facilitates seamless data sharing and utilization across platforms.

The two primary goals—achieving error-free semantic communication from a source to a target, and ensuring error-free generation of data required by the target based on the source data's meaning—are demonstrated to be logically and technically equivalent based on our developed LLM translation system.

### B. SEMANTIC INTEROPERABILITY BASED ON LLM "ADAPTER"

Semantic interoperability, the ability of different systems to understand and interpret data consistently, is crucial for seamless data communication and integration across diverse technological platforms. In a broader sense, the LLMs are integrated into a data processing pipeline and act as an interpretive layer. LLMs enable a more intuitive and accurate exchange of information between disparate data sources and systems.

### C. ENHANCING STANDARDS AND OPTIMIZATION THROUGH LLMs

Based on our experience during the experiment, the static data definition in ECLASS alone often falls short in covering all the dynamic needs for accurately annotating a given technical property. LLMs emerge as a powerful solution to this challenge, offering the ability to dynamically determine semantics and generate meaningful definitions of data.

Therefore, the relationship between AAS, ECLASS dictionary, and LLMs becomes synergistic. LLMs can facilitate the continuous improvement and adaptation of ECLASS dictionary library, ensuring they remain relevant and comprehensive to cover a large number of possible technical concepts. The dynamic semantic interpretation and generation capability of LLMs is also crucial for simplifying deployment and maintaining the utility of digital twins across various industrial applications, paving the way for a more interconnected and intelligent data exchange.

### D. DATA LIMITATION

The data used to test the effectiveness of this system include descriptive data of the technical properties of automation components (sensors, actuators, controllers, and network devices) selected from datasheets. These data are selected, because it is a typical usage scenario for participants in Industry 4.0 to share technical data model of components across different companies and the information systems [1], which has practical impacts and value. The generated AAS digital twin models in this work are limited to the technical data submodel according to AAS specifications. In the future, we believe the proposed method can be extended to other data types and usage scenarios (e.g., generating other AAS submodels standardized by Industry 4.0), and we anticipate similar results as evaluated in this study.

Additionally, incorporating LLMs to process various data sources within digital twins, such as simulation models [36], [37], offers the potential to expand semantic interoperability to provide predictive and operational insights. However, a significant challenge lies in serializing different knowledge representations into a textual format suitable for LLMs.

## VIII. CONCLUSION

This paper presents a comprehensive exploration into the automated generation of Asset Administration Shells using Large Language Models, with a focus on enhancing semantic interoperability within the framework of Industry 4.0. This approach facilitates the automated generation of AAS models to reduce human effort and implementation cost in creating AAS instances. By introducing a ''semantic node'' data structure to capture the essential semantics of a piece of information and developing a novel approach that leverages LLMs for the semantic translation of technical properties in AAS model, the system converts technical data into AAS models, which can be used for error-free information exchange in digital twins. Furthermore, the semantic definitions generated by the LLM system possess the capability to dynamically annotate information, and the resulting conceptual definition can be verified by human and be used to form a dictionary collection or enrich an external dictionary such as ECLASS.

The evaluation of various LLM configurations and the impact of the Retrieval-Augmented Generation (RAG) mechanism has yielded detailed insights into the deployment of LLM systems in this context. Our research demonstrates that LLMs can substantially streamline the AAS instance creation process, achieving up to a 78% pass rate in generating error-free and informative AAS model elements. This suggests that a significant portion of the effort typically required for crafting AAS models could be shifted toward more efficient validation processes. Open-source models deliver sufficient performance for this application and can be fairly used for commercial use. However, our comparative analysis indicates that the effectiveness of these models varies, depending on their acquired knowledge during training and the strategic integration of external semantic resources (RAG based on ECLASS dictionary). Based on our observations, we propose a new hypothesis regarding the ''cheat-sheet effect of RAG'' conjecture, which we plan to explore further. This paper serves as a proof of concept for using LLMs to automate the generation of AAS instances and proves the feasibility of using LLM as information interpreter for achieving semantic interoperability in digital twin systems in the context of Industry 4.0. The research result also lays the groundwork for further collaboration with research and industrial sectors to amplify our findings' value and practicality in the context of industrial digitalization and smart manufacturing.

## REFERENCES

[1] S. Bader, E. Barnstedt, H. Bedenbender, B. Berres, M. Billmann, and M. Ristin, ''Details of the asset administration shell—Part 1—The exchange of information between partners in the value chain of Industrie 4.0 (version 3.0 RC02),'' Plattform Ind. 4.0, Berlin, Germany, Tech. Rep. 3.0RC02, 2022.

[2] Y. Xia, N. Jazdi, and M. Weyrich, ''Automated generation of asset administration shell: A transfer learning approach with neural language model and semantic fingerprints,'' in *Proc. IEEE 27th Int. Conf. Emerg. Technol. Factory Autom. (ETFA)*, Sep. 2022, pp. 1–4, doi: 10.1109/ETFA52439.2022.9921637.

[3] F. Ocker, C. Urban, B. Vogel-Heuser, and C. Diedrich, ''Leveraging the asset administration shell for agent-based production systems,'' *IFAC-PapersOnLine*, vol. 54, no. 1, pp. 837–844, Jan. 2021, doi: 10.1016/j.ifacol.2021.08.186.

[4] G. Schnauffer, D. Görzig, C. Kosel, and J. Diemer, ''Asset administration shell for the wiring harness system,'' in *Advances in Automotive Production Technology—Towards Software-Defined Manufacturing and Resilient Supply Chains*, H. D. K. Niklas and Wulle, Eds. Cham, Switzerland: Springer, 2023, pp. 324–332.

[5] H. Eichelberger and C. Niederée, ''Asset administration shells, configuration, code generation: A power trio for Industry 4.0 platforms,'' in *Proc. IEEE 28th Int. Conf. Emerg. Technol. Factory Autom. (ETFA)*, Sep. 2023, pp. 1–8, doi: 10.1109/etfa54631.2023.10275339.

[6] N. Blasek, K. Eichenmüller, B. Ernst, N. Götz, B. Nast, and K. Sandkuhl, ''Large language models in requirements engineering for digital twins,'' in *Proc. 13th Enterprise Design Eng. Working Conf.*, Jun. 2023, pp. 1–15.

[7] C. Olah, N. Cammarata, L. Schubert, G. Goh, M. Petrov, and S. Carter, ''Zoom in: An introduction to circuits,'' *Distill*, vol. 5, no. 3, Mar. 2020, Art. no. e00024.001, doi: 10.23915/distill.00024.001.

[8] W. Gurnee, N. Nanda, M. Pauly, K. Harvey, D. Troitskii, and D. Bertsimas, ''Finding neurons in a haystack: Case studies with sparse probing,'' 2023, *arXiv:2305.01610*.

[9] T. Miny, M. Thies, U. Epple, and C. Diedrich, ''Model transformation for asset administration shells,'' in *Proc. IECON 46th Annu. Conf. IEEE Ind. Electron. Soc.*, Oct. 2020, pp. 2207–2212, doi: 10.1109/IECON43393.2020.9254649.

[10] M. Platenius-Mohr, S. Malakuti, S. Grüner, J. Schmitt, and T. Goldschmidt, "File- and API-based interoperability of digital twins by model transformation: An IIoT case study using asset administration shell," *Future Gener. Comput. Syst.*, vol. 113, pp. 94–105, Dec. 2020, doi: 10.1016/j.future.2020.07.004.
[11] M. Platenius-Mohr, S. Malakuti, S. Grüner, and T. Goldschmidt, "Interoperable digital twins in IIoT systems by transformation of information models: A case study with asset administration shell," in *Proc. 9th Int. Conf. Internet Things*. New York, NY, USA: Association for Computing Machinery, Oct. 2019, doi: 10.1145/3365871.3365873.
[12] J. Zhao, B. Vogel-Heuser, F. Bi, J. Höfgen, F. Ocker, B. Vojanec, T. Markert, and A. Kraft, "A semi-automatic approach for asset administration shell creation from heterogeneous data," *IFAC-PapersOnLine*, vol. 56, no. 2, pp. 3673–3679, 2023, doi: 10.1016/j.ifacol.2023.10.1532.
[13] L. M. V. Da Silva, A. Köcher, M. S. Gill, M. Weiss, and A. Fay, "Toward a mapping of capability and skill models using asset administration shells and ontologies," in *Proc. IEEE 28th Int. Conf. Emerg. Technol. Factory Autom. (ETFA)*, Sep. 2023, pp. 1–4, doi: 10.1109/etfa54631.2023.10275459.
[14] Y. Huang, S. Dhouib, L. P. Medinacelli, and J. Malenfant, "Enabling semantic interoperability of asset administration shells through an ontology-based modeling method," in *Proc. 25th Int. Conf. Model Driven Eng. Languages Syst.* New York, NY, USA: Association for Computing Machinery, 2022, pp. 497–502, doi: 10.1145/3550356.3561606.
[15] C. Schmidt, F. Volz, L. Stojanovic, and G. Sutschet, "Increasing interoperability between digital twin standards and specifications: Transformation of DTDL to AAS," *Sensors*, vol. 23, no. 18, p. 7742, Sep. 2023, doi: 10.3390/s23187742.
[16] N. Braunisch, M. Ristin-Kaufmann, R. Lehmann, M. Wollschlaeger, and H. W. van de Venn, "Generation of digital twins for information exchange between partners in the Industrie 4.0 value chain," in *Proc. IEEE 21st Int. Conf. Ind. Informat. (INDIN)*, Jul. 2023, pp. 1–6, doi: 10.1109/indin51400.2023.10218306.
[17] S. Rongen, N. Nikolova, and M. van der Pas, "Modelling with AAS and RDF in Industry 4.0," *Comput. Ind.*, vol. 148, Jun. 2023, Art. no. 103910, doi: 10.1016/j.compind.2023.103910.
[18] Y. Huang, S. Dhouib, L. P. Medinacelli, and J. Malenfant, "Semantic interoperability of digital twins: Ontology-based capability checking in AAS modeling framework," in *Proc. IEEE 6th Int. Conf. Ind. Cyber-Physical Syst. (ICPS)*, May 2023, pp. 1–8, doi: 10.1109/ICPS58381.2023.10128003.
[19] J. Fuchs, J. Schmidt, J. Franke, K. Rehman, M. Sauer, and S. Karnouskos, "I4.0-compliant integration of assets utilizing the asset administration shell," in *Proc. 24th IEEE Int. Conf. Emerg. Technol. Factory Autom. (ETFA)*, Sep. 2019, pp. 1243–1247, doi: 10.1109/ETFA.2019.8869255.
[20] T. Moreno, T. Sobral, A. Almeida, A. L. Soares, and A. Azevedo, "Semantic asset administration shell towards a cognitive digital twin," in *Flexible Automation and Intelligent Manufacturing: Establishing Bridges for More Sustainable Manufacturing Systems*, M. A. S. Francisco, Ed. Cham, Switzerland: Springer, 2024, pp. 679–686.
[21] A. Lüder, A.-K. Behnert, F. Rinker, and S. Biffl, "Generating Industry 4.0 asset administration shells with data from engineering data logistics," in *Proc. 25th IEEE Int. Conf. Emerging Technol. Factory Automat. (ETFA)*, Sep. 2020, pp. 867–874, doi: 10.1109/ETFA46521.2020.9212149.
[22] S. Cavalieri and M. G. Salafia, "Insights into mapping solutions based on OPC UA information model applied to the Industry 4.0 asset administration shell," *Computers*, vol. 9, no. 2, p. 28, Apr. 2020, doi: 10.3390/computers9020028.
[23] J. Beermann, R. Benfer, M. Both, J. Müller, and C. Diedrich, "Comparison of different natural language processing models to achieve semantic interoperability of heterogeneous asset administration shells," in *Proc. IEEE 21st Int. Conf. Ind. Informat. (INDIN)*, Jul. 2023, pp. 1–6, doi: 10.1109/indin51400.2023.10218154.
[24] M. Both, J. Müller, and C. Diedrich, "Automated mapping of semantically heterogeneous I4.0 asset administration shells by methods of natural language processing," *At-Automatisierungstechnik*, vol. 69, no. 11, pp. 940–951, Nov. 2021, doi: 10.1515/AUTO-2021-0050.
[25] A. Cartus, M. Both, M. Nicolai, J. Müller, and C. Diedrich, "Interoperability of semantically heterogeneous digital twins through natural language processing methods," in *Proc. CLIMA*, May 2022, pp. 1–8, doi: 10.34641/clima.2022.143.
[26] M. Both, B. Kämper, A. Cartus, J. Beermann, T. Fessler, D. J. Müller, and D. C. Diedrich, "Automated monitoring applications for existing buildings through natural language processing based semantic mapping of operational data and creation of digital twins," *Energy Buildings*, vol. 300, Dec. 2023, Art. no. 113635, doi: 10.1016/j.enbuild.2023.113635.
[27] A. Radford, R. Jozefowicz, and I. Sutskever, "Learning to generate reviews and discovering sentiment," 2017, *arXiv:1704.01444*.
[28] A. Vaswani, N. Shazeer, N. Parmar, J. Uszkoreit, L. Jones, A. N. Gomez, L. Kaiser, and I. Polosukhin, "Attention is all you need," 2017, *arXiv:1706.03762*.
[29] A. Rogers, O. Kovaleva, and A. Rumshisky, "A primer in BERTology: What we know about how BERT works," *Trans. Assoc. Comput. Linguistics*, vol. 8, pp. 842–866, Dec. 2020, doi: 10.1162/tacl_a_00349.
[30] S. Bills. (2023). *Language Models Can Explain Neurons in Language Models*. Accessed: May 14, 2023. [Online]. Available: https://openaipublic.blob.core.windows.net/neuron-explainer/paperindex.html
[31] D. Dittler, P. Lierhammer, D. Braun, T. Müller, N. Jazdi, and M. Weyrich, "A novel model adaption approach for intelligent digital twins of modular production systems," in *Proc. IEEE 28th Int. Conf. Emerg. Technol. Factory Autom. (ETFA)*, Sep. 2023, pp. 1–8, doi: 10.1109/etfa54631.2023.10275384.
[32] Y. Xia, M. Shenoy, N. Jazdi, and M. Weyrich, "Towards autonomous system: Flexible modular production system enhanced with large language model agents," in *Proc. IEEE 28th Int. Conf. Emerg. Technol. Factory Automat. (ETFA)*, Apr. 2023, pp. 1–8, doi: 10.1109/ETFA54631.2023.10275362.
[33] H. Touvron, L. Martin, K. Stone, P. Albert, A. Almahairi, Y. Babaei, N. Bashlykov, S. Batra, P. Bhargava, S. Bhosale, and D. Bikel, "Llama 2: Open foundation and fine-tuned chat models," 2023, *arXiv:2307.09288*.
[34] A. Q. Jiang, A. Sablayrolles, A. Roux, A. Mensch, B. Savary, C. Bamford, D. S. Chaplot, D. D. L. Casas, E. B. Hanna, F. Bressand, and G. Lengyel, "Mixtral of experts," 2024, *arXiv:2401.04088*.
[35] C. Xu, Q. Sun, K. Zheng, X. Geng, P. Zhao, J. Feng, C. Tao, and D. Jiang, "WizardLM: Empowering large language models to follow complex instructions," 2023, *arXiv:2304.12244*.
[36] P. Liu, J. Xing, Y. Li, C. Miller, and P. Tang, "Knowledge sharing and workforce engagement using digital twins-based simulations and extended reality for process operations," in *Proc. Comput. Civil Eng.*, Jan. 2024, pp. 680–687, doi: 10.1061/9780784485231.081.
[37] Y. Xia, D. Dittler, N. Jazdi, H. Chen, and M. Weyrich, "LLM experiments with simulation: Large language model multi-agent system for process simulation parametrization in digital twins," 2024, *arXiv:2405.18092*.

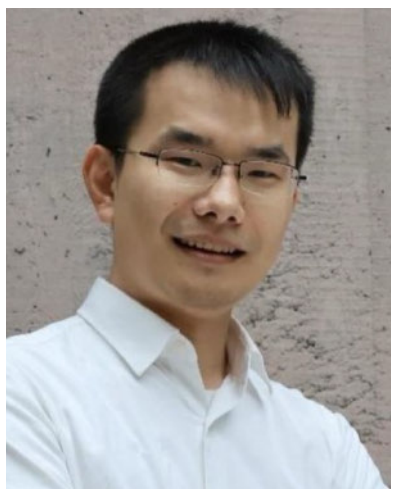

**YUCHEN XIA** (Member, IEEE) received the dual bachelor's degrees in mechanical design and automation from Wuhan University, China, and in automotive and engine technology from the University of Stuttgart, Germany, in 2017, and the master's degree from the University of Stuttgart in 2019, with the specialization in automotive mechatronics and automated driving.

In 2019 and 2020, he was with the Fraunhofer-Institut für Produktionstechnik und Automatisierung (Fraunhofer IPA), as a Research Assistant. In 2020, his research proposal received financial backing from Stiftung der Deutschen Wirtschaft, which aims to support entrepreneurship. Since 2021, he has been conducting his doctoral research, as an Academic Researcher with the Institute of Industrial Automation and Software Engineering (IAS), University of Stuttgart. His research interests include digital twins in industrial automation, large language models, multi-agent systems, and industrial autonomous systems.

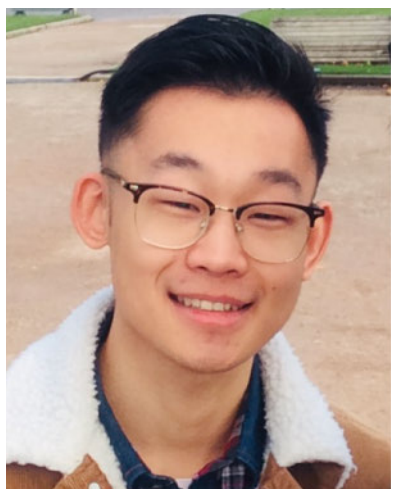

**ZHEWEN XIAO** (Member, IEEE) received the M.S. degree in electrical engineering and information technology from the University of Stuttgart, Germany, in 2023. He was a graduate student at the University of Stuttgart and will continue his academic career in teaching and research at the School of Information and Electronic Engineering, Hunan City University, China. His research interests include industrial automation, multimodal large models, and digital twins.

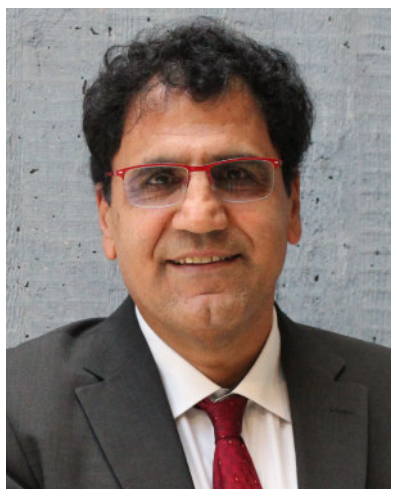

**NASSER JAZDI** (Senior Member, IEEE) received the Diploma degree in electrical engineering and the Ph.D. degree in remote diagnosis and maintenance of embedded systems from the University of Stuttgart, Germany, in 1997 and 2003, respectively.

In 2003, he joined the Institute of Industrial Automation and Software Engineering (IAS), University of Stuttgart. He gives two lectures: "software engineering" and "reliability and security of automation systems." In 2009, he worked for two months as an Invited Researcher with Prof. Zadeh with the University of California at Berkeley. He is currently the Deputy Head and the Academic Director of the Institute of Industrial Automation and Software Engineering, University of Stuttgart, and a Visiting Professor with Anhui University. His research interests include software reliability in the context of the IoT, learning aptitude for industrial automation, and artificial intelligence in industrial automation. He is a member of the VDE Association for Electrical, Electronic and Information Technologies, the VDI-GPP Software Reliability Group, and Berkeley Initiative in Soft Computing (BISC).

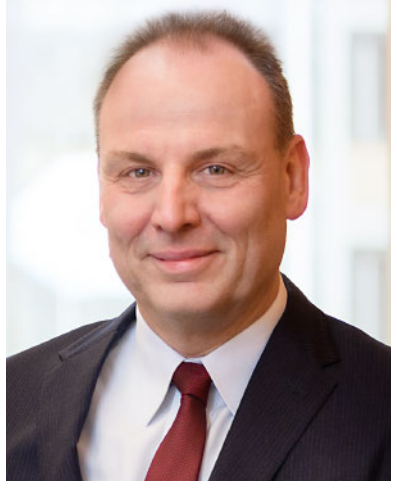

**MICHAEL WEYRICH** (Member, IEEE) was born in Saarbrücken, Germany, in 1967. He received the Dipl.-Ing. degree in automation technology from the Hochschule für Technik und Wirtschaft des Saarlandes, Saarbrücken, in 1991, and the Dr.-Ing. degree in mechanical engineering from RWTH Aachen University, Germany, in 1999.

His major field of expertise was industrial automation. He has spent over ten years in the industry, including several years abroad, primarily in roles related to automation technology and software systems for manufacturing. He has been the Head of the Institute for Automation Technology and Software Engineering (IAS), University of Stuttgart, Germany, since 2013. He is currently with European Center for Mechatronics, RWTH Aachen University, and held significant positions with Daimler AG and Siemens AG, focusing on CAx process chains for production and motion control/PLM software development. His research interests include intelligent automation systems, the safety of AI-based systems, and complexity management in automation technology. He is the Chairperson of the Board of Directors of the VDI/VDE Society for Measurement and Automation Technology, from 2022 to 2025. He was honored with an Honorary Doctorate (Dr. h.c.) from Donetsk National Technical University, in 2018, for his scientific contributions. He also been actively involved in various committees and editorial boards related to automation technology and factory automation.